\documentclass[10pt,twocolumn,letterpaper]{article}
\usepackage{cvpr}              

\usepackage{graphicx}
\usepackage{amsmath}
\usepackage{amssymb}
\usepackage{booktabs}

\usepackage{multirow}

\usepackage[ruled,vlined]{algorithm2e}

\renewcommand\footnotemark{}

\usepackage[pagebackref,breaklinks,colorlinks]{hyperref}

\usepackage{titling}
\date{}
\predate{}
\postdate{}
\usepackage[resetlabels]{multibib}
\newcites{Main}{References}%
\newcites{Supp}{References}%

\usepackage[capitalize]{cleveref}
\crefname{section}{Sec.}{Secs.}
\Crefname{section}{Section}{Sections}
\Crefname{table}{Table}{Tables}
\crefname{table}{Tab.}{Tabs.}

\def\cvprPaperID{2150} 
\def\confName{CVPR}
\def\confYear{2023}

\begin{document}

\title{\vspace{-0.2em} \Large \textbf{Token Boosting for Robust Self-Supervised Visual Transformer Pre-training}   \vspace{0.3em}}

\author{Tianjiao Li\textsuperscript{1\dag}
~~~ Lin Geng Foo\textsuperscript{1\dag}\thanks{\dag~equal contribution}
~~~ Ping Hu\textsuperscript{2}
~~~ Xindi Shang\textsuperscript{3} 
~~~ Hossein Rahmani\textsuperscript{4}\\
Zehuan Yuan\textsuperscript{3}
~~~ Jun Liu\textsuperscript{1\ddag}\thanks{\ddag~corresponding author}\\
\textsuperscript{1}Singapore University of Technology and Design\\
\textsuperscript{2}Boston University ~~ \textsuperscript{3}ByteDance ~~ \textsuperscript{4}Lancaster University\\
{\tt\small \{tianjiao\_li,lingeng\_foo\}@mymail.sutd.edu.sg, pinghu@bu.edu, shangxindi@bytedance.com
 } \\
{\tt\small h.rahmani@lancaster.ac.uk, yuanzehuan@bytedance.com, jun\_liu@sutd.edu.sg }
}

\maketitle

\begin{abstract}
Learning with large-scale unlabeled data has become a powerful tool for pre-training Visual Transformers (VTs). However, prior works tend to overlook that, in real-world scenarios, the input data may be corrupted and unreliable. Pre-training VTs on such corrupted data can be challenging, especially when we pre-train via the masked autoencoding approach, where both the inputs and masked ``ground truth" targets can potentially be unreliable in this case. To address this limitation, we introduce the Token Boosting Module (TBM) as a plug-and-play component for VTs that effectively allows the VT to learn to extract clean and robust features during masked autoencoding pre-training. We provide theoretical analysis to show how TBM improves model pre-training with more robust and generalizable representations, thus benefiting downstream tasks. We conduct extensive experiments to analyze TBM's effectiveness, and results on four corrupted datasets demonstrate that TBM consistently improves performance on downstream tasks.
\end{abstract}

\vspace{-0.4cm}

\section{Introduction}

Having rapidly risen in popularity in recent years, Vision Transformer (ViT) \cite{dosovitskiy2021an} and its variants \cite{han2022survey} have shown impressive performance across various computer vision tasks, such as image classification, video recognition and 3D action analysis \cite{bao2022beit,he2021masked,wang2022bevt,shi2020decoupled,li2022dynamic}.
Amongst this wave of research on Visual Transformers (VTs), there has emerged a popular paradigm -- self-supervised VT pre-training \cite{bao2022beit,he2021masked,caron2021dino,wang2022bevt} -- which has attracted a lot of attention in the research community.
These works \cite{bao2022beit,he2021masked,caron2021dino,wang2022bevt} generally pre-train VTs on a large dataset in a self-supervised manner, allowing them to extract semantically meaningful and generalizable features without the need for annotated data.
These pre-trained VTs are practical and effective, showing good downstream performance with only slight fine-tuning, and have quickly become an important research direction.

Among the various self-supervised pre-training strategies, masked autoencoding \cite{bao2022beit,he2021masked,dosovitskiy2021an} is a prominent approach that has been widely explored.
Masked autoencoding \cite{bao2022beit,he2021masked,dosovitskiy2021an,wang2022beit3,wang2022bevt,chen2022hierarchically,yu2022pointbert,liang2022meshmae} works by randomly masking a portion of input tokens or patches, and letting the VT reconstruct them.
Thus, in order to successfully reconstruct the masked tokens or patches, the VT is driven to learn the underlying structure and semantic information of the input data.
Due to its natural compatibility with the token-wise representation in VTs, masked autoencoding has been explored for pre-training VTs on data across many fields, 
such as RGB images \cite{bao2022beit,he2021masked,wang2022beit3}, pose data \cite{chen2022hierarchically,liang2022meshmae} and 3D data \cite{yu2022pointbert}.

\begin{figure}[t]
    \centering
    \includegraphics[width=0.8\linewidth]{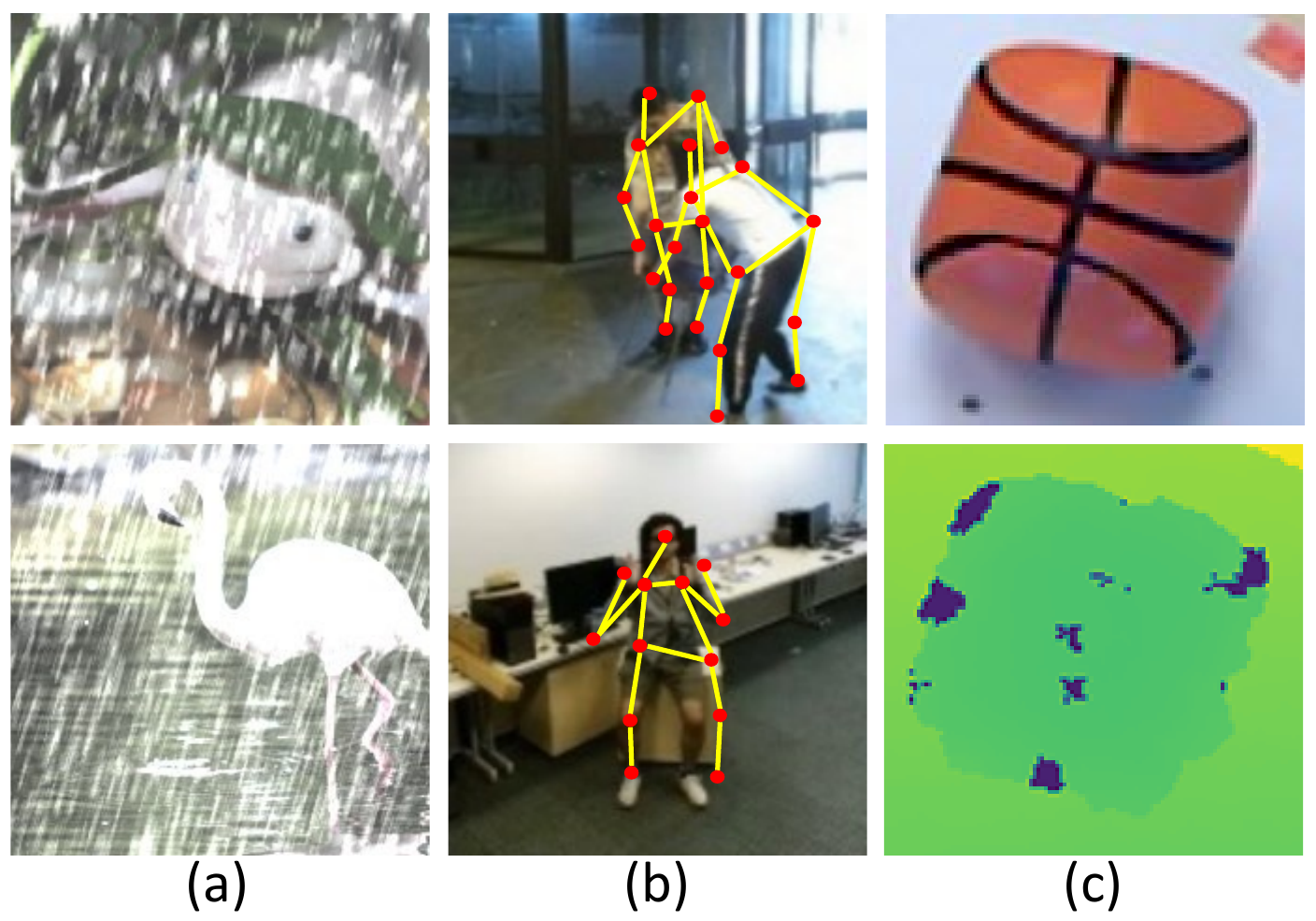}
    \vspace{-0.3cm}
    \caption{
    Samples of corrupted data. 
    (a) Images in adverse weather conditions.
    (b) Skeletons with corresponding RGB images. Skeleton data collected with a pose tracking algorithm \cite{wang2015evaluation} can often contain misplaced joints. 
    (c) Depth image (Bottom) with its corresponding RGB image (Top). The collected depth image can be unreliable, e.g., having noticeable perturbations.
    Images are taken from ImageNet-C \cite{hendrycks2019robustness}, NTU RGB+D 120 \cite{liu2019ntu} and Washington RGB-D \cite{lai2011large} datasets.
    Best viewed in color. 
    }
    \label{fig:noisy_samples}
    \vspace{-0.5cm}
\end{figure}

However, in many real-world scenarios, the input data can be of low quality and can be unreliable, which is often overlooked.
For example, in adverse weather conditions \cite{hnewa2020object}, the quality of captured images can be quite bad, with many corrupted pixels that degrade the image.
Depth images, which are commonly used for 3D analysis, also often contain errors and noticeable perturbations due to measurement errors introduced by the depth camera sensors \cite{lai2011large}.
Skeleton data can also have distortions in the joint coordinates \cite{liu2016spatio,demisse2018pose,wang2015evaluation} due to noise introduced by the sensors, e.g., the pose tracking function of Kinect v2 often introduces average offsets of 50-100 mm per joint \cite{wang2015evaluation}. 
Examples of such corrupted data are depicted in Fig.~\ref{fig:noisy_samples}.

In many cases, we have access to large-scale unlabeled data containing such corruptions and noise, and would like to exploit them to pre-train a VT that can be effectively adapted to a wide range of downstream tasks \cite{wang2022beit3,he2021masked,bao2022beit,chen2022hierarchically}.
Yet, this can be rather challenging since the data corruptions can interfere with the learning of representations during self-supervised pre-training, leading to unreliable features and predictions in downstream tasks. This is the case in particular for masked autoencoding approaches, where both the inputs and the ``ground truth" targets can be corrupted at the same time during pre-training.
Hence, we aim to tackle the key question:
How can we improve the VT's ability to extract robust and reliable features during self-supervised pre-training on corrupted data?
Once this is achieved, the pre-trained capability can then be transferred to downstream tasks, where the VTs can reliably extract generalizable features 
and perform well even when the testing data is unreliable.

To this end, we propose a novel Token Boosting Module (TBM) to improve the robustness of VTs against unreliable 
and corrupted data during the pre-training process.
TBM is a plug-and-play module that can improve the performance of VTs by boosting the reliability of tokens, making them cleaner such that we can feed robust features to the subsequent layers.
We also provide analysis to theoretically demonstrate that our TBM can indeed encourage the learning of cleaner features and tokens during masked autoencoding pre-training.

In summary, our contributions are:
\begin{itemize}

    \item     
    We design a novel Token Boosting Module (TBM) to learn to extract reliable features from corrupted and unreliable data during the self-supervised pre-training process of VTs. The TBM can be easily incorporated into multiple layers of VTs, and trained in an end-to-end manner to improve robustness and performance on corrupted data.    

    \item We provide theoretical analysis to show that the robustness capabilities of TBM can be learned during the self-supervised masked autoencoding pre-training.    
    
    \item We conduct experiments on multiple tasks including RGB image classification, 3D skeleton action recognition, and depth image classification with corrupted data. Experiments consistently show that our TBM provides significant improvements in downstream tasks.
\end{itemize}

\section{Related Work}

\noindent{\textbf{Self-supervised pre-training.}}
As a practical way to train models without the need for annotated data, self-supervised pre-training has received a lot of attention. 
Recently, masked autoencoding has become a very popular approach for self-supervised Visual Transformer (VT) pre-training \cite{dosovitskiy2021an,bao2022beit,he2021masked,chen2020generative,peng2022beitv2,wang2022beit3,wang2022bevt,liu2022masked}, together with other alternatives
including contrastive learning \cite{radford2021clip,jia2021align,yuan2021florence,yu2022coca,chen2021mocov3,chen2021simsiam,zhang2022morphmlp} 
and self-distillation \cite{caron2021dino,grill2020byol,li2021albef}. 
Due to the popularity of pre-training methods and their potential widespread applications in real-world scenarios \cite{bao2022beit,he2021masked,chen2022hierarchically,liang2022meshmae,yu2022pointbert,liu2022masked}, it becomes important to improve their robustness to corrupted data, but this is often overlooked by previous methods.
For instance, during masked autoencoding pre-training on corrupted data, both inputs and ``ground truths" can be unreliable, making it a tricky situation.
Therefore, we explore if, under these challenging circumstances, we can allow VTs to learn to extract robust feature representations.

\noindent{\textbf{Robustness of VTs.}}
Visual Transformers have recently been a popular and hot research direction, and their robustness has also attracted attention.
Some existing works 
\cite{bhojanapalli2021understanding,bai2021transformers,zhou2022understanding,mao2022towards} 
investigate the test-time robustness of VTs, while some others explore the adversarial robustness of VTs 
\cite{mahmood2021robustness,mao2022towards,jiang2020robust}.
Different from these works, we aim to improve the VTs' capability to extract clean and robust features during self-supervised pre-training. 
This capability can benefit downstream tasks and improve their robustness against corrupted input data at test time.

\noindent{\textbf{Handling of Corrupted Data.}}
Improving deep models’ performance on corrupted data is important for many practical tasks~\cite{liu2016spatio,demisse2018pose,moran2020noisier2noise,shi2019probabilistic,carlucci20182,zaki2019viewpoint}.
Previous efforts have been made to handle corrupted input data for supervised settings in tasks like skeleton action recognition \cite{liu2016spatio,demisse2018pose,chen2021ctrgcn,yan2018spatial,shi2019two,cheng2020skeleton}, depth image classification \cite{carlucci20182,zaki2019viewpoint}, and face recognition \cite{shi2019probabilistic}.
Alternatively, some other approaches \cite{Yan2018ddrnet,Sterzentsenko2019self,moran2020noisier2noise,xie2020noise2same} can be performed during data pre-processing to reconstruct a ``clean" input sample before using them for corresponding tasks.
Many of these methods \cite{Yan2018ddrnet,Sterzentsenko2019self,moran2020noisier2noise,xie2020noise2same,shahroudy2016ntu} require an additional stage of training, and tend to be specially designed for specific types of data or corruptions only.
Here, we investigate a different problem from the above works, and instead explore the handling of corrupted data during \textit{self-supervised VT pre-training}.

\section{Method}

In many real-world scenarios, the data (e.g., depth images, images in adverse weather conditions) can be collected with substantial corruptions.
When such corrupted data is used for self-supervised pre-training, they can lead to unreliable features and predictions, interfering with the learning of generalizable and robust features.
In order to improve VTs' performance on corrupted and unreliable data, we introduce a Token Boosting Module (TBM) for VTs that can learn to boost the reliability of features during self-supervised pre-training.

TBM is a plug-and-play module that can be placed between layers in the VT, and can be trained in an end-to-end manner. 
When placed in the intermediate layers, TBM can make features in VTs cleaner and more reliable.
To further improve performance, we incorporate TBM into multiple layers of the VT, such that contextual information at multiple feature levels are used for boosting the quality of the features.
This enhanced capability of VTs, achieved during self-supervised pre-training, can thus be transferred to downstream tasks to achieve good performance even when facing corrupted inputs.

In this section, we first give a brief background of masked autoencoding. Then, we explain the feature boosting technique and describe TBM in detail. Lastly, we theoretically show why our TBM can benefit pre-trained VTs with more robust and reliable representations.

\subsection{Masked Autoencoding Pre-training}
Previous works \cite{dosovitskiy2021an,bao2022beit,he2021masked} have explored masked autoencoding to train Transformers, which we follow.
In masked autoencoding pre-training \cite{dosovitskiy2021an,bao2022beit,he2021masked}, a part of the input is often masked, and the Transformer (with a decoder head that is only used during pre-training) is tasked to reconstruct those masked inputs.
Driven by the reconstruction losses, the Transformer will learn to capture semantically meaningful information within the input, producing features that tend to generalize well to other downstream tasks. 
However, this masked autoencoding approach can be challenging when performed with corrupted data, since both the input and masked target could be corrupted, impairing downstream tasks with biased/unreliable feature representations.

\subsection{Feature Boosting Technique}
\label{section:theoretical_motivations}

In this subsection, inspired by previous works \cite{moran2020noisier2noise,xu2020noisyasclean}, we describe a simple technique to extract robust and reliable features from data corrupted by noise.
Specifically, for an unreliable feature $F$ that is going to be fed into a VT layer, this technique allows us to obtain a reliable boosted feature $\hat{R}$.
The robust feature $\hat{R}$ (instead of the unreliable feature $F$) can then be fed into the VT layer for better performance.

Let $R$ represent the ground truth reliable features.
We are unable to directly observe the reliable feature $R$,
but are able to observe the unreliable feature $F$ (which are extracted from a real input sample). 
Importantly, this means that we are unable to simply train a module with direct supervision to produce $R$ from $F$.
We assume that the difference between $R$ and $F$ can be modelled by a feature perturbation $P$ which is randomly drawn from a distribution $\mathcal{P}$, i.e., we have $F = R + P$.
Below, we show how, with only access to the unreliable feature $F$, we can produce $\hat{R}$ (which statistically speaking is an unbiased conditional expectation \cite{steyer2017probability,speyer2008stochastic}) that estimates the reliable feature $R$.

\begin{figure}
    \centering
    \includegraphics[width=1\linewidth]{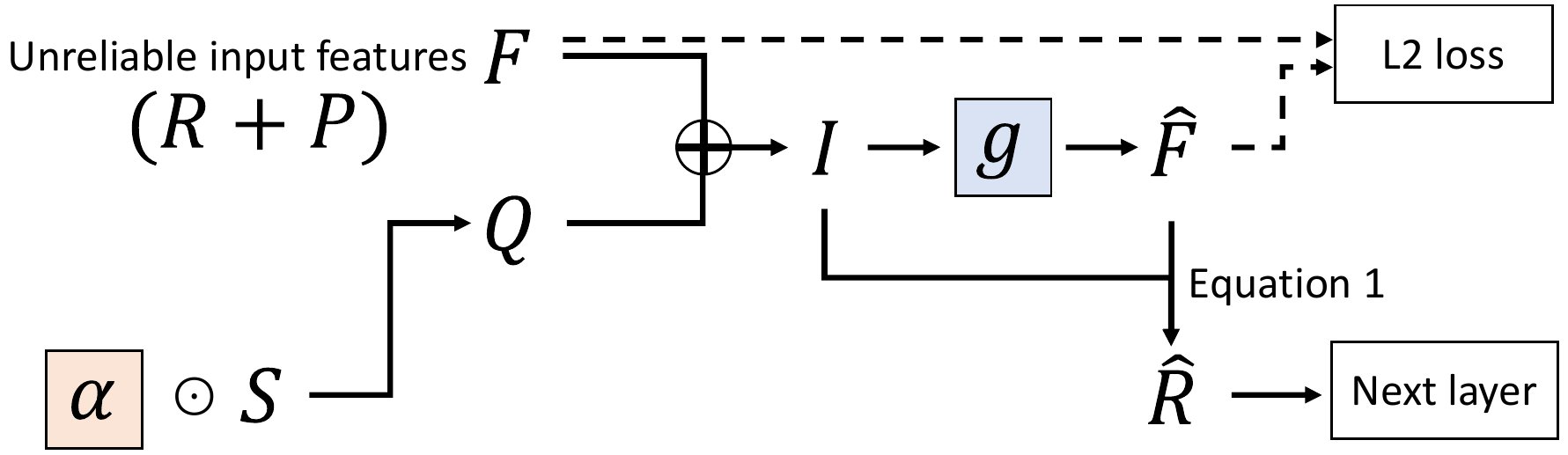}
    \vspace{-0.35cm}
    \caption{
    Illustration of the TBM placed at a VT layer.
    Firstly, TBM takes in the unreliable input feature $F$, i.e., all tokens.
    Synthetic corruption $Q$ is produced by scaling Gaussian noise $S$ with learned parameters $\alpha$.
    An intermediate representation $I=F+Q$ is produced, and fed into the autoencoder $g$ to reconstruct $\hat{F}$, an estimate of $F$.    
    Lastly, we apply Eq.~\ref{eqn:xhat} with $\hat{F}$ and $I$ as input, producing $\hat{R}$,
    which is our boosted output feature.
    $\hat{R}$ is then fed as input to the next VT layer.
    }
    \label{fig:TBM}
    \vspace{-0.35cm}
\end{figure}

Given $F$, we first create an intermediate representation, $I = F + Q$ where 
$Q$ is a synthetic corruption drawn from $\mathcal{P}$ as well, as shown in Fig. \ref{fig:TBM}.
Importantly, as $P$ and $Q$ are i.i.d., the two sources of corruptions in $I$ will be indistinguishable to an observer, i.e., we will not be able to tell how much of the corruption comes from $P$ or $Q$.
We then train an autoencoder $g$ (parameterized by $\theta$) to predict $F$ given $I$ as its input, and we denote the output of $g$ as $\hat{F}$ (an estimate of $F$), as shown in Fig. \ref{fig:TBM}. $\hat{F}$ will later be used to construct the boosted features $\hat{R}$.

Specifically, the autoencoder $g$ is trained with an L2 loss with $F$ as the target, and will be optimized to reconstruct good estimates of $F$ given $I$ as input,
i.e., $\hat{F} = \mathbb{E}[F |I]$.
Notably, in order to reproduce $F$ perfectly, the autoencoder needs to distinguish between $P$ and $Q$, but since it never observes $P$ and $Q$ independently, it cannot reproduce $F$ fully, and instead can only produce its best estimate $\hat{F}$.
Expanding $\hat{F}$, we observe that $\hat{F} = \mathbb{E}[F |I] = \mathbb{E}[R + P |I] = \mathbb{E}[R|I] + \mathbb{E}[P|I]$.

Next, as $Q$ and $P$ are assumed to be independently drawn from the same distribution, we get $\mathbb{E} [Q | I] =  \mathbb{E} [P| I]$. 
Intuitively, this can be understood as $P$ and $Q$ being indistinguishable from one another, and a simple proof has been provided in the Supplementary. 
Multiplying the above equation by $2$ and performing substitution, we get
$2\mathbb{E}[F |I] = \mathbb{E}[R|I] + (\mathbb{E}[R|I] + \mathbb{E}[P|I] + \mathbb{E}[Q|I]) = \mathbb{E}[R|I] + \mathbb{E}[R + P +Q|I] = \mathbb{E}[R|I] + \mathbb{E}[F +Q|I] = \mathbb{E}[R|I] + I$.

By rearranging, we get $\mathbb{E}[R|I] = 2\mathbb{E}[F |I] - I$. 
We can thus obtain $\hat{R}$  using an estimate $\hat{F}$ as:
\vspace{-0.12cm}
\begin{equation}
    \label{eqn:xhat}
    \hat{R} = 2 \hat{F} - I.
\end{equation}

In summary, by using this technique, we can still estimate the reliable feature $\hat{R}$ even if we do not have ground truth feature $R$. 
This is important, because we do not have access to any ground truth clean data or features ($R$).

\subsection{Token Boosting Module}

Next, we present the Token Boosting Module (TBM) to improve  reliability of VTs by incorporating the technique described above.
One practical issue we face, is how to sample synthetic corruption $Q$, which requires us to make assumptions on the distribution $\mathcal{P}$.
Previous methods \cite{shi2019probabilistic,li2019unsupervised,yu2019robust,lee2018simple} handling corrupted data show that modeling perturbations in the deep feature space as Gaussian distributed can be very effective for learning to approximate complicated and intractable distributions underlying the data.
Following them, we apply the Gaussian distribution to model the feature perturbation $P\sim\mathcal{P}$, with a mean of $0$ and unknown standard deviations.
Notably, the Gaussian assumption \cite{shi2019probabilistic,li2019unsupervised,yu2019robust,lee2018simple} allows us to ``parameterize" corruptions in feature space, and we can then optimize the distribution of the added synthetic corruption $Q$ to be similar to the underlying corruption distribution $\mathcal{P}$, that fulfills the conditions in Sec.~\ref{section:theoretical_motivations}. We empirically observe this works effectively for handling various types of corruptions and tasks.

Let the input feature of TBM be $F \in \mathbb{R}^K$.
As shown in Fig.~\ref{fig:TBM}, our TBM consists of two sets of learnable parameters.
Firstly, we have the scaling parameters $\alpha \in \mathbb{R}_{\geq 0}^K$ that scale the added corruptions.
This is necessary to learn the \textit{optimal standard deviation parameters of the synthetic corruptions}, where a different standard deviation is learned for each of $K$ feature dimensions.
Importantly, our training loss will optimize $\alpha$ towards making the distribution similar to the natural corruption's distribution $\mathcal{P}$, therefore \textit{the synthetic corruptions and the natural corruptions can be seen as i.i.d.}\footnote{We provide detailed analysis that distribution of $Q$ can be trained to approximate distribution of $P$ in Sec.~5 of Supplementary.}. 
Secondly, we have the autoencoder $g$ containing learnable parameters $\theta$. 
This autoencoder $g$ takes in $F$ with added corruptions $Q$, and reconstructs $\hat{F}$, an estimate of $F$.

We show an illustration of the TBM in Fig.~\ref{fig:TBM}. 
First, we sample a Gaussian noise vector $S \in \mathbb{R}^K$ with a mean of $0$ and standard deviation of $1$.
The sampled Gaussian noise $S$ is scaled using the learned parameters $\alpha$ in an element-wise manner (denoted by $\odot$). 
Note that $\alpha \odot S$ corresponds to sampling $K$ Gaussians, with the standard deviation of the Gaussian at each index $k$ to be $\alpha_k$.
We use $\alpha \odot S$ as synthetic corruption $Q$, and add it to $F$ to form $I$ as follows:
\vspace{-0.2cm}
\begin{equation}
    I = F + (\alpha \odot S).
\end{equation}

Next, the autoencoder $g$ is used to reconstruct the estimate of $F$ from $I$:
\begin{equation}
    \hat{F} = g (I).
\end{equation}
\vspace{-0.5cm}

Then, we utilize Eq.~\ref{eqn:xhat} to obtain a boosted feature $\hat{R}$, which is used as the input to the next layer of the VT.

\paragraph{TBM for features.}
We insert TBM between self-attention layers in VTs, such that the tokens at the feature layers can be boosted before generating the next set of key, query and value embeddings.
In order to further improve the robustness of VTs, we incorporate TBM into multiple layers of the VT.
This allows us to enhance the intermediate features of VTs in the presence of different levels of contextual information, which leads to better robustness capabilities.

\paragraph{L2 Reconstruction Loss.}
To train the autoencoder parameters $\theta$, we introduce an L2 reconstruction loss $L_{recon}$:
\begin{equation}
    L_{recon}(F,\hat{F}) = \lambda \sum_{k=1}^K [F_k - \hat{F}_k]^2
\end{equation}
where $\lambda$ is a hyperparameter that controls the weightage of the loss with respect to the end-to-end pre-training objective.
This loss $L_{recon}$ trains the autoencoder $g$ to produce good estimates $\hat{F}$, and can be combined with the end-to-end objective in masked autoencoding pre-training to drive TBM parameters $\theta$ and $\alpha$ towards improved robustness capabilities, as described in the next sub-section.

\subsection{TBM in Masked Autoencoding Pre-training}
Our main goal in this paper is to improve the capability of VTs to extract robust features on corrupted and unreliable data during masked autoencoding pre-training, which is a popular and effective paradigm for pre-training VTs.
However, one key question remains: \textit{can the TBM be meaningfully trained under the masked autoencoding pre-training objective?}
This is not immediately straightforward, especially since both the inputs and the reconstruction targets are unreliable.
In this section, we answer the above question and theoretically show how $\alpha$ and $\theta$ can be meaningfully trained with the masked autoencoding objective.

We first show our theoretical motivations in more detail using a linear example for the sake of simplicity.
Specifically, we explain how a simple linear regression model, tasked to regress some masked inputs $V \in \mathbb{R}^{N_V}$ from the unmasked inputs $U \in \mathbb{R}^{N_U}$, will potentially be able to \textit{achieve a lower loss from the use of our TBM}. This shows that \textit{gradients from the loss will lead to meaningful training of the TBM parameters}.
In this example, we assume that the best-fit linear relationship between $U$ and $V$ can be modelled as $V = \beta U + c + \epsilon$, where $\beta \in \mathbb{R}^{N_V \times N_U},c \in \mathbb{R}^{N_V},\epsilon \in \mathbb{R}^{N_V}$ respectively refer to the coefficient, intercept and random error coming from a Gaussian with standard deviation of $\gamma$. 
We additionally assume that at least some elements in $\beta$ are non-zero, which means that there is at least some correlation between $U$ and $V$.

When trained with adequate data, the fitted linear model (comprising of parameter estimates $\hat{\beta}$ and $\hat{c}$) will converge to $\beta$ and $c$, with the MSE error as follows:
\begin{equation}
\label{eq:mse_clean}
\mathbb{E} \big[ \frac{1}{N_V} \sum_{j=1}^{N_V} [V_j- (\beta U + c)_j]^2  \big]  = \frac{1}{N_V} \sum_{j=1}^{N_V}  \mathbb{E} [\epsilon_j^2] =  \gamma^2,
\end{equation}
where the detailed steps are shown in Eq.~23-28 of Supplementary.
This $(\gamma^2)$ represents the best possible result that the linear model can achieve.
However, when ``natural" corruptions are added towards all inputs in $U$ and $V$ to get $\tilde{U}$ and $\tilde{V}$ (where $\tilde{U} = U + S_U$ and $\tilde{V} = V + S_V$, with $S_V$ and $S_U$ containing elements drawn from a Gaussian with standard deviation $\sigma_N$), the MSE will be increased to:
\begin{align}
    \mathbb{E}  &\bigg[ \frac{1}{N_V}  \sum_{j=1}^{N_V}  [\tilde{V}_j - (\beta \tilde{U} + c)_j]^2 \bigg] \\
    &= \frac{1}{N_V} \sum_{j=1}^{N_V}  \mathbb{E} [\epsilon_j^2] + \mathbb{E} [(S_V)_j^2] +  \mathbb{E} [(\beta S_U)_j^2 ] \\
    \label{eq:mse_both_corrupted}
    &= \gamma^2 + \sigma_N^2 + \frac{1}{N_V} \sum_{j=1}^{N_V} \sum_{k=1}^{N_U} \beta_{jk}^2 \sigma_N^2.
\end{align}

Refer to Eq.~29-40 of Supplementary for more details of the above derivations.
The increase in MSE from Eq.~\ref{eq:mse_clean} to Eq.~\ref{eq:mse_both_corrupted} represents the extra difficulty in conducting inference, when there are corruptions in the input data $U$ (contributing $\frac{1}{N_V} \sum_{j=1}^{N_V} \sum_{k=1}^{N_U} \beta_{jk}^2 \sigma_N^2$), as well as corruptions in the ``ground truth" inputs $V$ (contributing $\sigma_N^2$).
We can thus potentially reduce the MSE to $\gamma^2 + \sigma_N^2$ by boosting $\tilde{U}$  (which represents the corrupted inputs) and feeding to the model $\hat{U} \approx U$, which our TBM, if placed before our linear regression model, \textit{will be optimized to do}.
In other words, if we can boost $\tilde{U}$ into a clean version $\hat{U} \approx U$, our expected loss (with the full proof in Eq.~41-50 of Supplementary) will be:
\begin{align}
    \mathbb{E}  \bigg[ \frac{1}{N_V} & \sum_{j=1}^{N_V} [\tilde{V}_j - (\beta U + c)_j]^2  \bigg] \\
    &= \frac{1}{N_V} \sum_{j=1}^{N_V}  \mathbb{E} [\epsilon_j^2]  + \mathbb{E} [(S_V)_j^2] \\
    &= \gamma^2 + \sigma_N^2,
\end{align}
which is lower than the MSE if $\tilde{U}$ is directly used (shown in Eq. \ref{eq:mse_both_corrupted}).
Thus, during the optimization process, when our TBM is updated via the loss gradients that aim to minimize the pre-training loss, our TBM will tend to learn feature boosting capabilities in order to achieve a smaller loss.
This enables the feature boosting capabilities (i.e. learned parameters $\alpha$ and $\theta$) of our TBM to be trained, such that it can function as described in Sec. \ref{section:theoretical_motivations}.

Although the above is just a linear example, we remark that the concepts involved can generalize to non-linear models such as VTs.
In general, \textit{even for non-linear cases}, unreliable inputs will lead to difficulties in masked patch prediction, which will lead to a higher pre-training loss than reliable inputs.
Thus, in order to minimize this loss, the gradients will optimize TBM parameters $\alpha$ and $\theta$ towards learning how to boost the features, i.e., produce good estimates $\hat{R} = R$,
which the TBM's design as described in Sec.~\ref{section:theoretical_motivations} explicitly allows it to accomplish.
Our experimental results also demonstrate the efficacy of such a design.

Note that when trained with supervision labels, TBM can also be trained to boost the reliability of features as well, as input corruptions are likely to negatively affect performance of the supervised task (e.g. 3D action recognition \cite{shi2020decoupled,chen2021ctrgcn,liu2020disentangling,foo2023unified}, depth image classification \cite{qiao2021private,zaki2019viewpoint,loghmani2019recurrent}) such that the gradients for $\alpha$ and $\theta$ will tend to encourage the learning of feature boosting capabilities.

\section{Experiments}

We first conduct extensive experiments on
image classification with added synthetic corruptions, which allows us to systematically vary the type of added corruptions for a more detailed analysis.
Then, we test our method on two real-world settings, namely 3D action recognition and depth image classification, where we attain good performance.
More experiments can be found in Supplementary.

\subsection{Training Setups}
In order to evaluate our method more comprehensively, we conduct experiments with two different training setups: the \textbf{Self-supervised Training} setup and the \textbf{Supervised Training} setup, which we explain below.

\paragraph{Self-supervised Training.} Following previous works \cite{he2021masked,dosovitskiy2021an,bao2022beit}, we add a decoder on top of our VT encoder (inserted with some TBM modules), then
we mask input patches and train our VT encoder+decoder by reconstructing these masked patches using MSE loss.
After the end-to-end self-supervised training, we follow the \textit{linear probe} setting \cite{he2021masked} to train a classifier for testing purposes. 
Specifically, the decoder is replaced by a classification head, and we train the classification head while keeping the parameters in our VT encoder (including our TBMs) frozen.
Then, we report the results using the predictions from the VT encoder+classification head.

\paragraph{Supervised Training.} We also aim to evaluate the capability of TBM when trained in a supervised setting.
Therefore, we train our VT encoder (including the TBM modules), in a normal supervised manner, where a classification head is attached to the end of the VT encoder and the full model is trained in an end-to-end manner. The classification head is reused at test time for evaluation.

\subsection{Image Classification }

\paragraph{Dataset.}
We conduct both self-supervised and supervised training on ImageNet \cite{deng2009imagenet} with the synthetic corruptions from ImageNet-C \cite{hendrycks2019robustness}.
ImageNet-C is a corrupted version of the ImageNet test set, and is a commonly used benchmark for robustness evaluation, employing 15 types of synthetically generated corruptions, and some examples can be seen in Fig.~\ref{fig:noisy_samples}.
ImageNet-C allows us to comprehensively and systematically evaluate the effectiveness of TBM on various types of common corruptions.
Specifically, during both \textit{supervised} and \textit{self-supervised training}, we follow the procedure for ImageNet-C \cite{hendrycks2019robustness} and add synthetic corruptions for 50\% of input images, where the corruption types are randomly selected (among the 15 types). 
During testing, we evaluate our method on the corrupted ImageNet-C \cite{hendrycks2019robustness} benchmark.

\noindent{\textbf{Network Architecture.}}
To comprehensively evaluate the efficacy of the proposed TBM, we perform experiments using three different sizes of ViT \cite{dosovitskiy2021an}: ViT-Huge, ViT-Large and ViT-Base, as well as DeiT \cite{touvron2021training} and Swin Transformer (Swin) \cite{liu2021swin}. We use the Base versions of DeiT and Swin.
Following previous works \cite{he2021masked,bao2022beit}, the pre-training decoders are composed of stacked transformer blocks.
The autoencoder $g$ in our proposed TBM is comprised of 3 fully-connected+ReLU layers with hidden layers having dimensionality $input\_dim / 2$, where $input\_dim$ depends on the architecture of the VT.
Scaling values $\alpha$ have the same shape as the input feature map $F$ to the TBM, which depends on the architecture.
Our TBM is inserted at 3 different feature levels, i.e., bottom layer, middle layer, and top layer in our ViT, DeiT and Swin backbones, in order to enhance feature robustness with different levels of contextual information.

\noindent \textbf{Training Details.} In the \textit{self-supervised training} phase, we follow existing works \cite{he2021masked} and use the AdamW optimizer to train the VTs (for all ViTs, DeiT and Swin) for $400$ epochs, where $75\%$ of input patches are randomly masked. The initial learning rate is set to $0.01$ and gradually decay to $0$ following the cosine rule. 
In the \textit{supervised training} phase, we follow \cite{he2021masked} to use LARS optimizer for ViT and we use AdamW optimizer for DeiT and Swin. All experiments are conducted on Nvidia A100 GPUs, and $\lambda$ is set to $1$.

\noindent \textbf{Results.} 
Our results are shown in Tab.~\ref{tab:imagenetc}. 
Overall, our TBM provides significant performance gains for all five baseline VTs in both self-supervised and fully-supervised settings. 
This shows that TBM is capable of improving the learning on corrupted data.

\begin{table}[ht]
\footnotesize
   \centering
    \caption{Performance comparison (\%) over multiple baseline VTs on Image Classification using ImageNet-C. 
    }
    \vspace{-0.3cm}
       \begin{tabular}{ccc} 
	   \toprule
       Methods & Supervised & Self-supervised \\
	   \midrule
 	   ViT-Huge \cite{dosovitskiy2021an} & 50.1 & 35.2\\
	   ViT-Huge+TBM  & \textbf{53.2}  & \textbf{38.6} \\
       \midrule
       ViT-Large \cite{dosovitskiy2021an} & 48.7  & 33.4 \\
       ViT-Large+TBM &  \textbf{51.5}  &  \textbf{36.1} \\
       \midrule
       ViT-Base \cite{dosovitskiy2021an} & 41.6 & 25.9   \\
       ViT-Base+TBM & \textbf{45.2} & \textbf{29.3}  \\
	   \midrule
       DeiT \cite{touvron2021training} & 44.2 & 28.7   \\
       DeiT+TBM & \textbf{47.7} & \textbf{32.1}  \\	   
	   \midrule
       Swin \cite{liu2021swin} & 45.1 &  29.6 \\
       Swin+TBM & \textbf{47.9} & \textbf{33.6}  \\	          
      \bottomrule
      \end{tabular}
	\label{tab:imagenetc} 
\vspace{-0.15cm}
\end{table}

Next, we conduct more in-depth analysis using ViT-Huge as the backbone model.
We evaluate the effects of TBM on various types of corruptions in Tab.~\ref{tab:ablation_different_types}.
We observe that TBM improves performance consistently across \textit{all corruption types} (noise, blue, weather, digital) used in ImageNet-C.
We remark that this includes a 3.1\% increase on images corrupted with snow, and a 3.6\% increase on images corrupted with motion blur, which are examples of practical yet complicated corruptions to tackle.
These results suggest that TBM has the capability to generalize to various types of corruptions.
We also conduct experiments on another setting \textbf{ViT-Huge+TBM (Individual)}, where a single model is trained to deal with each individual type of corruption.
This setting leads to a slight improvement as expected, since this setting allows our TBM to fully specialize in tackling a single type of corruption at a time.
Nevertheless, our significant improvements over the baseline show that our method (which boosts the features at multiple levels using TBM) is capable of handling these various synthetic corruptions simultaneously, which is a complicated task.

\begin{table*}[ht]
\footnotesize
\caption{
Performance comparison (\%) of TBM on different types of corruptions on ImageNet-C. 
} 
\vspace{-0.3cm}
\centering
\begin{tabular}{l p{0.34cm} p{0.34cm} p{0.34cm} | p{0.34cm} p{0.34cm} p{0.5cm} p{0.5cm} | p{0.34cm} p{0.34cm} p{0.34cm} p{0.5cm} | p{0.5cm} p{0.5cm} p{0.34cm} p{0.34cm} c}
\hline
\multirow{2}{*}{Methods} &  \multicolumn{3}{c}{Noise} & \multicolumn{4}{c}{Blur} & \multicolumn{4}{c}{Weather} & \multicolumn{4}{c}{Digital} & \multirow{2}{*}{Mean}\\
\cline{2-16}
& \scriptsize Gauss. & \scriptsize Shot & \scriptsize Imp. & \scriptsize Defoc. & \scriptsize Glass & \scriptsize Motion & \scriptsize Zoom & \scriptsize Snow & \scriptsize Frost & \scriptsize Fog & \scriptsize Bright & \scriptsize Contrast & \scriptsize Elastic & \scriptsize Pixel & \scriptsize JPEG  \\
\hline
ViT-Huge \cite{dosovitskiy2021an} & 36.7 & 35.6 & 35.0 & 34.3 & 28.5 & 39.2 & 34.4 & 34.6 & 26.7 & 23.8 & 45.8 & 28.6 & 35.9 & 42.8 & 45.6 & 35.2 \\
ViT-Huge+TBM & 39.5 & 39.2 & 38.7 & 37.6 & 30.4 & 42.8 & 37.2 & 37.7 & 29.5 & 26.9 & 49.8 & 32.2 & 39.6 & 47.1 & 49.2 & 38.6 \\
ViT-Huge+TBM (individual) & 39.8 & 39.4 & 39.1 & 37.9 & 30.7 & 42.9 & 37.3 & 38.0 & 29.9 & 27.2 & 50.1 & 32.6 & 40.0 & 47.4 & 49.9  & 38.8 \\
\bottomrule
\end{tabular}
\label{tab:ablation_different_types} 
\vspace{-0.2cm}
\end{table*}

\noindent \textbf{Evaluation with data pre-processing methods.}
We also want to explore if data pre-processing techniques can be applied to effectively improve the quality of the data before performing self-supervised pre-training.
We evaluate the performance of some common smoothing methods, namely the \textbf{Median Filter} and the \textbf{Gaussian Filter}, as well as three image-based pre-processing methods.
As shown in Tab.~\ref{tab:ablation_denoising}, we find that performance improvements are small.
This can be because there are many different types of corruptions, and it can be difficult for a single pre-processing technique to tackle them effectively. 
It can be more effective to perform the boosting and ``cleaning" at the feature level, like our TBM does.
Furthermore, we also explore using these five data pre-processing techniques in conjunction with our method (\textbf{[Pre-processing] + ViT-Huge + TBM}), where we apply these methods to the input image before feeding them into our ViT-Huge+TBM model.
The results when we use these pre-processing techniques in conjunction with our method are comparable to directly using our TBM, suggesting that it is sufficient to directly use our TBM, which indicates the efficacy of our method.

\begin{table}[t]
\footnotesize
   \centering
    \caption{
    Performance comparison (\%) with data pre-processing methods on ImageNet-C.
    }
    \vspace{-0.3cm}
  \resizebox{0.473\textwidth}{!}{
  \begin{tabular}{@{}ccc@{}}    
	   \toprule
       Methods & Supervised & Self-supervised \\
	   \midrule
 	   ViT-Huge \cite{dosovitskiy2021an} & 50.1 & 35.2\\
 	   \midrule
 	   Median Filter & 51.3 &  36.3\\
 	   Gaussian Filter & 51.0 &  36.1 \\
       Shi \etal \cite{shi2019probabilistic} & 51.8 & 36.4 \\
       Moran \etal \cite{moran2020noisier2noise} & 52.0 & 36.6  \\ 
       Xie \etal \cite{xie2020noise2same}  & 51.9 & 36.5 \\    
 	   \midrule
	   Median Filter + ViT-Huge + TBM  & 52.9  & 38.5\\
	   Gaussian Filter + ViT-Huge + TBM &  52.7  & 38.2\\
	   Shi \etal \cite{shi2019probabilistic} + ViT-Huge + TBM & 53.0 & 38.5 \\
       Moran \etal \cite{moran2020noisier2noise} + ViT-Huge + TBM & 53.1 & 38.7   \\ 	   	   
       Xie \etal \cite{xie2020noise2same} + ViT-Huge + TBM  & 53.0 & 38.6 \\    
 	   \midrule	   
	   ViT-Huge+TBM  & 53.2  & 38.6 \\
      \bottomrule
      \end{tabular}
	\label{tab:ablation_denoising} 
}
\end{table}

\noindent \textbf{Layers to add TBM.}
In our method, we apply TBM across several layers spanning from the bottom to the top of the VT.
Here, we ablate this decision by evaluating several alternative designs on ImageNet-C, with results shown in Tab.~\ref{tab:ablation_TBM}.
\textbf{Top layer}, \textbf{Mid layer} and \textbf{Bot layer} are settings where the TBM is only applied to the top layer, the middle layer, or the bottom layer.
Compared to using it in any single layer, our incorporation into multiple layers performs better, demonstrating the efficacy of exploiting context information at different feature levels.

\begin{table}[h]
\footnotesize
\caption{
Evaluation of TBM applied to different layers.
We also conduct testing with TBM incorporated into more layers, and do not find further improvement.
} 
\centering
\vspace{-0.3cm}
\begin{tabular}{l | c c c c c c }
\toprule
\multirow{2}{*}{Settings}  & \multirow{2}{*}{Baseline \cite{dosovitskiy2021an}} &  Top & Mid  & Bot  &  Ours  \\
         &  &  layer & layer  & layer  &  (All 3 layers) \\
\midrule
Acc. (\%) & 35.2 & 37.2 & 37.1 & 37.3  & 38.6 \\
\bottomrule
\end{tabular}
\label{tab:ablation_TBM} 
\end{table}

\noindent \textbf{Weightage of losses.}
We conduct ablation studies on the impact of $\lambda$ in Tab.~\ref{tab:ablation_lambda}. $\lambda = 1$ performs the best. 
When $\lambda$ is too low, the TBM is not trained well, and when $\lambda$ is too high, the VT may lose focus on the main objective. 
Specifically, when we set $\lambda=0$, we do not train with the reconstruction loss, which effectively disables our boosting mechanism, leading to sub-optimal performance.
Notably, when $\lambda = 0$, the model also contains the same number of model parameters as our main ViT-Huge+TBM setting reported in Tab.~\ref{tab:imagenetc}, but performs significantly worse, which indicates that the benefits of our TBM does not merely come from the slightly ($\sim 5\%$) increased number of parameters.

\begin{table}[ht]
\small
\caption{
Evaluation of different values of $\lambda$ for TBM on ImageNet-C.
} 
\centering
\vspace{-0.3cm}
\begin{tabular}{l | c c c c c c}
\toprule
$\lambda$ & 0 & 0.1 & 0.5 & 1 &  2 &  5\\
\midrule
Acc. (\%) & 35.7 & 37.4 & 38.1 & 38.6 & 38.5 & 38.5 \\
\bottomrule
\end{tabular}
\label{tab:ablation_lambda} 
\end{table}

\noindent \textbf{Evaluation on clean ImageNet.}
We also evaluate the accuracy of ViT-Huge+TBM to be 85.3\% on the clean test set of ImageNet, while the baseline ViT-Huge achieves an accuracy of 85.1\%.
As accuracy remains about the same as the original ViT-Huge, we conclude that the benefits of TBM mainly lie in improved robustness against corrupted data.

\subsection{3D Skeleton Action Recognition}

In 3D skeleton action recognition, models are tasked to classify the performed action using 3D skeleton sequences as input. 
However, there are often errors in the 3D skeleton data \cite{liu2016spatio}, which will negatively affect performance.
Thus, this is a real-world setting where we can evaluate the effectiveness of our TBM.

\noindent \textbf{Dataset.} NTU RGB+D 60 (NTU60) \cite{shahroudy2016ntu} is a large dataset that has been widely used for 3D action recognition, containing approximately 56k RGB-D sequences from 60 activity classes collected with Kinect v2 sensors.
NTU RGB+D 120 (NTU120) \cite{liu2019ntu} is an extension of NTU60, and is currently the largest RGB+D dataset for 3D action analysis, containing more than 114k skeletal sequences across 120 activity classes.
We follow the standard evaluation protocols of \cite{shahroudy2016ntu,liu2019ntu} for 3D action recognition,
where the Cross-Subject (xsub) and Cross-View (xview) protocols are tested for NTU60, while the Cross-Subject (xsub) and Cross-Setup (xset) protocols are tested for NTU120.

\noindent \textbf{Network Architecture.}
We perform experiments using DSTA-Net \cite{shi2020decoupled}, which is specifically designed for processing skeleton data, and currently achieves state-of-the-art performance among VTs in this field.
Following previous works \cite{degardin2022generative,yu2020structure}, we design the decoder (used during pre-training) as a stack of GCN blocks.
Our TBM settings here are similar to the image classification experiments, where autoencoder $g$ consists of 3 fully-connected+ReLU layers, and TBM is inserted at 3 different feature levels, from low to high.

\noindent  \textbf{Pre-training Settings.} 
For masked autoencoding pre-training for 3D sequences, we randomly mask ``spatial" patches as well as ``temporal" patches, and task DSTA-Net to reconstruct them.
Thus, DSTA-Net will be trained to encode information about both the \textit{spatial aspect of the human body} and the \textit{temporal aspect of motions}, which will lead to its features capturing generalizable information that benefits downstream tasks. 
More details are in Supplementary.

\noindent \textbf{Training Details.}
In the self-supervised training phase, we train our model for 200 epochs with batch size of 256 using the AdamW optimizer. We set the initial learning rate to $0.01$, and set it to gradually decay to $0$ following the cosine rule.
$\lambda$ is set to $1$.
In the \textit{supervised training} phase, we use SGD optimizer following \cite{shi2020decoupled}.

\noindent \textbf{Results on Self-Supervised Training.} We compare with the current state-of-the-art self-supervised models in Tab.~\ref{tab:action_recognition_selfsupervised}.
It can be observed that adding our TBM to DSTA-Net and then conducting pre-training improves the performance of DSTA-Net consistently across all settings.
This also includes a significant 6.0\% and 4.0\% improvement over DSTA-Net baselines on NTU60 xsub and xview respectively.
We also highlight that DSTA-Net+TBM achieves good results on all reported settings.

\begin{table}[h]
\footnotesize
    \centering
    \makeatletter\def\@captype{table}\makeatother\caption{\small{
    Performance comparison (\%) on 3D Skeleton Action Recognition. 
    We follow the evaluation setting of~\cite{shahroudy2016ntu,liu2019ntu}. Models are pre-trained in a self-supervised manner, and tested with a linear probe. We attain good performance on all settings. 
    }
    }
	\label{tab:action_recognition_selfsupervised} 
	\vspace{-0.3cm}
	\centering
    \begin{tabular}{c  c c  c c }
    \toprule
    \multirow{2}{*}{Methods} &  \multicolumn{2}{c}{NTU60} & \multicolumn{2}{c}{NTU120} \\
    \cline{2-3} \cline{4-5}
     & xsub & xview & xsub & xset \\
    \midrule 
    LongT GAN \cite{zheng2018unsupervised} & 39.1 & 52.1 & - & - \\
    M$S^2$L \cite{lin2020ms2l} & 52.6 & -  & - & - \\
    P\&C FS-AEC \cite{su2020predict} & 50.6 & 76.3 & - & - \\
    P\&C Fw-AEC \cite{su2020predict} & 50.7 & 76.1 & - & - \\
    SeBiReNet \cite{nie2021view} & - & 79.7 & - & 69.3 \\
    EnGAN-PoseRNN \cite{kundu2019unsupervised}  & 68.6  & 77.8 & -  &  - \\
    T-Colorization \cite{yang2021skeleton} & 71.6 & 79.9 & - & -  \\
    TS-Colorization \cite{yang2021skeleton} & 74.6 & 82.6 & - & -  \\
    TSP-Colorization \cite{yang2021skeleton} & 75.2 & 83.1 & - &  - \\
    SkeletonCLR \cite{li20213d} & 68.3 & 76.4 & 56.8 & 55.9 \\
    3s-SkeletonCLR \cite{li20213d} & 75.0 & 79.8 & 60.7 & 62.6 \\
    CroSCLR \cite{li20213d} & 68.3 & 76.4 & 56.8 & 55.9 \\
    3s-CroSCLR \cite{li20213d} & 77.8 & 83.4 & - & -\\ 
    \midrule
    DSTA-Net \cite{shi2020decoupled} & 73.1 & 81.5 & 68.6 & 70.1  \\
    DSTA-Net+TBM & \textbf{79.1} & \textbf{85.5} & \textbf{69.7}& \textbf{71.1} \\
    \bottomrule
    \end{tabular}
\end{table}

\begin{table}[ht]
\footnotesize
        \centering
        \caption{
        Performance comparison (\%) on 3D Skeleton Action Recognition with models trained in a fully-supervised manner.  
        We follow the  setting of \cite{shahroudy2016ntu,liu2019ntu}.
        }
    	\label{tab:action_recognition} 
    	\vspace{-0.3cm}
    	\begin{tabular}{c  c c  c c }
        \toprule
        \multirow{2}{*}{Methods} &  \multicolumn{2}{c}{NTU60} & \multicolumn{2}{c}{NTU120} \\
        \cline{2-3} \cline{4-5}
         & xsub & xview & xsub & xset \\
        \midrule 
        ST-LSTM+TG \cite{liu2016spatio} & 69.2  & 77.7 & - & - \\         
        ST-GCN \cite{yan2018spatial} & 81.5 & 88.3 & 70.7 & 73.2 \\
        AS-GCN \cite{li2019actional} & 86.8 & 94.2 & 78.3 & 79.8 \\
        2s-AGCN \cite{shi2019two}  & 88.5 & 95.1 & 82.2 & 84.1  \\
        DGNN \cite{shi2019skeleton} & 89.9 & 96.1 &  -  &  -  \\
        SGN \cite{zhang2020semantics} & 89.0 & 94.5  & 79.2 & 81.5  \\
        Shift-GCN \cite{cheng2020skeleton} & 90.7 & 96.5 & 85.9 & 87.6 \\  
        MS-G3D \cite{liu2020disentangling} & 91.5 & 96.2 & 86.9 & 88.4 \\        
        FGCN \cite{yang2021feedback} & 90.2 & 96.3 & 85.4 & 87.4 \\
        CTR-GCN \cite{chen2021ctrgcn} & 92.4 & 96.8 & 88.9 & 90.6  \\       
        \midrule
        DSTA-Net \cite{shi2020decoupled} & 91.5 & 96.4 & 86.6 & 89.0 \\
        DSTA-Net+TBM & \textbf{93.1} & \textbf{97.0}  & \textbf{89.1}  & \textbf{91.1}  \\
        \bottomrule
        \end{tabular}

        \caption{Performance comparison (\%) on Depth Image Classification using WRGBD dataset based on depth data only.
        }
        \label{tab:wrgbd} 
    	\begin{tabular}{c  c c  c c}
        \toprule
        Methods &  \multicolumn{1}{c}{Supervised} & \multicolumn{1}{c}{Self-supervised}  \\
        \midrule
        CNN-RNN \cite{socher2012convolutional} & 78.9 & - \\
        DECO \cite{carlucci20182}  &  84.0  &- \\
        VGG\_f-RNN \cite{caglayan2018exploiting}  &  84.0  & - \\
        RCFusion \cite{loghmani2019recurrent}  &  85.9  & - \\
        HP-CNN \cite{zaki2019viewpoint}  &  85.0  &  -\\
        MMFLAN \cite{qiao2021private}  &  84.0  & - \\
        \midrule
        ViT-Huge \cite{dosovitskiy2021an}   & 82.3 & 69.7    \\
        ViT-Huge+TBM & 86.1 & 71.5 \\
        \bottomrule
        \end{tabular}

\vspace{-0.43cm}
\end{table}

\noindent \textbf{Results on Supervised Training.}
In Tab.~\ref{tab:action_recognition}, we show the results of training DSTA-Net in a supervised manner for 3D skeleton action recognition. 
Applying TBM to DSTA-Net provides improvement gains over the baseline DSTA-Net over all reported settings, including a 1.6\% gain on NTU60 xsub and 2.5\% gain on NTU120 xsub.
This allows our DSTA-Net+TBM model to achieve good results for all reported settings.

\subsection{Depth Image Classification}

In depth image classification, models are tasked to classify images using only the depth image as input (without RGB data). This is challenging as the depth cameras often induce corruptions in the data, which leads to noticeable perturbations and noise.
Note that here, we only use the depth image for classification.

\noindent \textbf{Dataset.}
The Washington RGB-D Object (WRGBD) dataset \cite{lai2011large} contains 300 common household objects from 51 different categories. These objects are filmed using RGB-D cameras from multiple angles.

\noindent \textbf{Implementation Details.} We perform experiments using ViT-Huge \cite{dosovitskiy2021an}, and follow the architecture and training details from our image classification experiments.

\noindent \textbf{Results.} We evaluate our method on depth image classification in Tab.~\ref{tab:wrgbd}.
TBM provides a 3.8\% improvement on the supervised setting and a 1.8\% improvement on the self-supervised setting.
This further verifies the efficacy of TBM at tackling tasks with corrupted data.

\vspace{-0.04cm}

\section{Conclusion}

In this work, we propose a novel TBM for robust self-supervised VT pre-training on corrupted data. 
We also theoretically motivate how TBM can learn its feature boosting capabilities using the masked autoencoding pre-training task.
When incorporated into VTs and trained in a self-supervised or fully-supervised manner, our TBM leads to performance improvements on corrupted data.

\footnotesize
\noindent
\textbf{Acknowledgments.}
This work is supported by MOE AcRF Tier 2 (Proposal ID: T2EP20222-0035), National Research Foundation Singapore under its AI Singapore Programme (AISG-100E-2020-065), and SUTD SKI Project (SKI 2021\_02\_06).
This work is also supported by TAILOR, a project funded by EU Horizon 2020 research and innovation programme under GA No 952215.

\normalsize

{\small
\bibliographystyle{ieee_fullname}
\bibliography{egbib}
}

\newpage

\setlength{\textheight}{8.875in}
\setlength{\textwidth}{6.875in}
\setlength{\columnsep}{0.3125in}
\setlength{\topmargin}{0in}
\setlength{\headheight}{0in}
\setlength{\headsep}{0in}
\setlength{\parindent}{1pc}
\setlength{\oddsidemargin}{-.304in}
\setlength{\evensidemargin}{-.304in}

\title{\vspace{-0.2em} \Large \textbf{Token Boosting for Robust Self-Supervised Visual Transformer Pre-training (Supplementary Material)} \vspace{0.3em} }

\author{Tianjiao Li\textsuperscript{1\dag}
~~~ Lin Geng Foo\textsuperscript{1\dag}
~~~ Ping Hu\textsuperscript{2}
~~~ Xindi Shang\textsuperscript{3} 
~~~ Hossein Rahmani\textsuperscript{4}\\
Zehuan Yuan\textsuperscript{3}
~~~ Jun Liu\textsuperscript{1\ddag}\\
\textsuperscript{1}Singapore University of Technology and Design\\
\textsuperscript{2}Boston University ~~ \textsuperscript{3}ByteDance ~~ \textsuperscript{4}Lancaster University\\
{\tt\small \{tianjiao\_li,lingeng\_foo\}@mymail.sutd.edu.sg, pinghu@bu.edu, shangxindi@bytedance.com
 } \\
{\tt\small h.rahmani@lancaster.ac.uk, yuanzehuan@bytedance.com, jun\_liu@sutd.edu.sg }
}

\maketitle

\setcounter{section}{0}
\setcounter{table}{0}
\setcounter{figure}{0}
\setcounter{equation}{0}

\section{More Experiments}

\begin{figure}
    \centering
    \includegraphics[width=\linewidth]{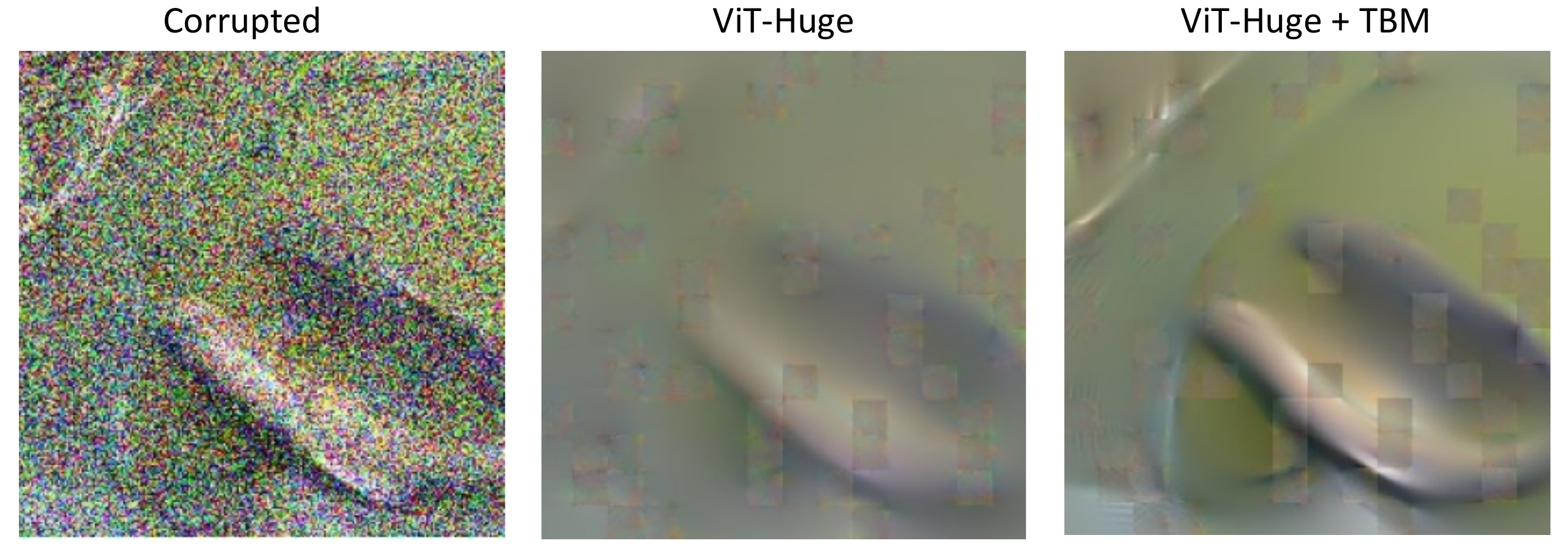}    
    \caption{  
    Visualization of corrupted image and its reconstruction during pre-training.        
    Here, we visualize a corrupted input image (left), the reconstruction from ViT-Huge (middle), and the reconstruction with token boosting from our ViT-Huge+TBM (right).    
    Best viewed in color.
    }
    \label{fig:vis_image}
\end{figure}

\textbf{Visualization of corrupted samples and features.}
Here, in order to qualitatively show the improvements from using TBM, we visualize the image reconstruction quality during pre-training.
In
Fig.~\ref{fig:vis_image}, 
we plot the original corrupted image and the reconstructed images of ViT-Huge with the decoder, with and without the use of TBM.
When TBM is not used, reconstruction of the masked parts of the image is challenging, and the reconstruction looks blurred and inaccurate.
This is because the corruptions in the unmasked parts of the image make it difficult to predict the masked parts.
However, when TBM is used, there is a visible improvement in the quality of the reconstructed images, where the image looks sharper and less blurred, and much of the corruptions have been smoothened out.

\begin{figure}
    \centering
    \includegraphics[width=\linewidth]{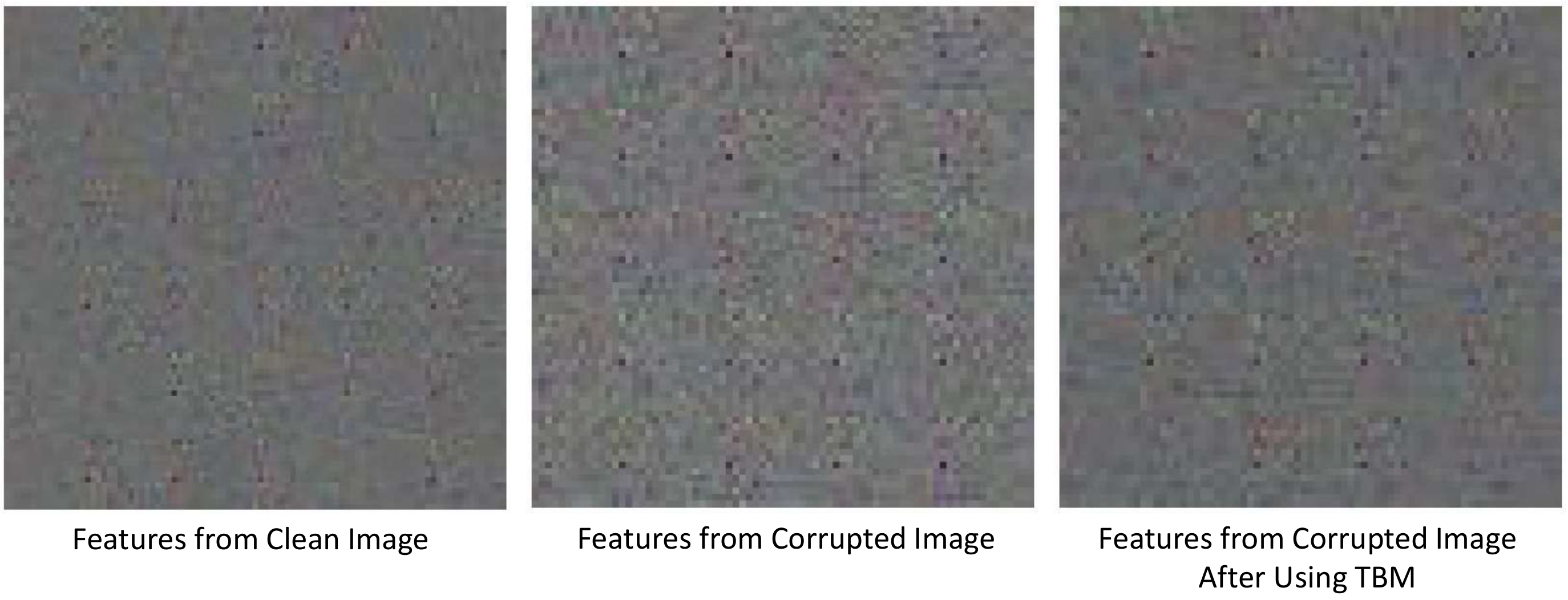}    
    \caption{
    Visualization of features.     
    We visualize the features obtained when a clean image is input to ViT-Huge (left), when a corrupted image is input to ViT-Huge (middle), and when the same corrupted image is input to ViT-Huge+TBM (right).       
    Best viewed in color.
    }
    \label{fig:vis_features}
\end{figure}

Next, in Fig.~\ref{fig:vis_features} 
we visualize the effects of corruptions on the extracted features. 
On the left, we see the produced features when the input image is clean.
When the same input is perturbed with a corruption and fed to the baseline ViT-Huge, the features undergo significant observable changes, showing that the features are not very robust to added corruptions.
However, when TBM is applied, the features show minimal changes when fed with the same  corrupted image, and look similar to features obtained under a clean setting.
This shows that TBM helps to make the output features of VTs more robust against input corruptions.

\begin{figure}
    \centering
    \includegraphics[width=\linewidth]{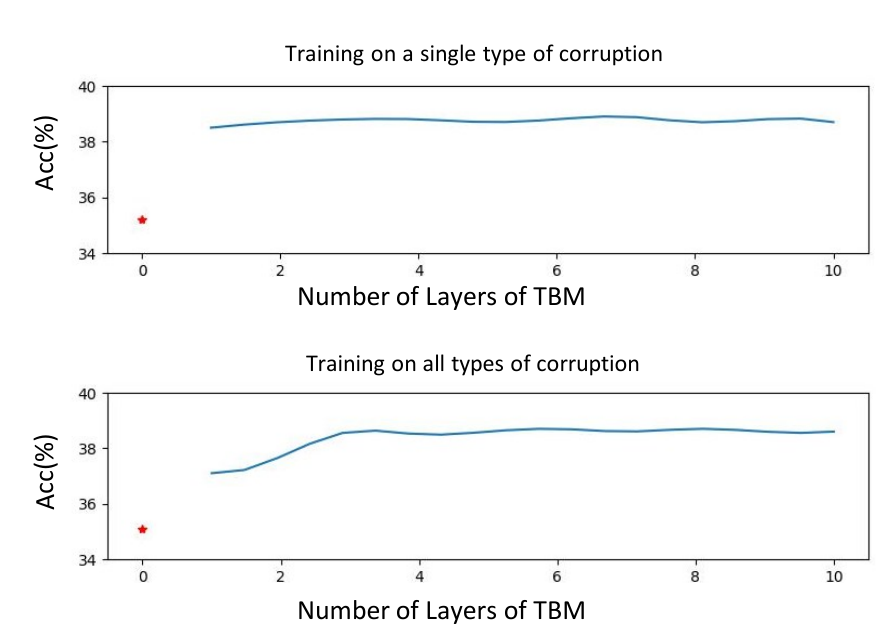}
    \caption{
    Evaluation of the impact of the number of layers using TBM modules.
    At the top, we plot the results where a single model is trained to deal with an individual type of corruption in ImageNet-C. At the bottom, we plot the results where the model is trained to handle all types of corruptions.
    The results of the baseline (i.e., ViT-Huge) without using any TBM modules are indicated in red stars (\textcolor{red}{$\star$}).
    }
    \label{fig:vis_num_layers}
\end{figure}

\textbf{Investigation on the impact of the number of layers using TBM modules.}
Firstly, we train a single model to deal with an individual type of
corruption in a self-supervised manner, i.e.,
we train multiple models (with ViT-Huge as the encoder) to handle the various types of corruptions, and report the average performance of the models over all the corruption types in the top row of Fig.~\ref{fig:vis_num_layers}.
We find that, after adding our TBM module to a single layer, we obtain a significantly improved accuracy.
However, inserting TBM modules to more layers does not lead to further improvement, which suggests that inserting our TBM module to a single layer is sufficient in this case. 
Next, we train one model to handle all types of corruptions in ImageNet-C at the same time.
The results are plotted in the bottom row of Fig.~\ref{fig:vis_num_layers}.
As the number of layers using our TBM module increases, the performance increases by a large margin and then becomes stable after 3 layers.
The results suggest that,
our model, which boosts the features at multiple levels (3 levels in our experiments) with TBM, is capable of handling various types of corruptions simultaneously, which is a complicated task.

\textbf{Performance after fine-tuning our pre-trained model.}
In the main paper, we evaluate on the self-supervised and supervised settings, and here we evaluate on a third setting, where we conduct fine-tuning on the entire model (i.e., both the ViT-Huge+TBM encoder and the linear layer).
Specifically, we fine-tune the model for 40 epochs using the AdamW optimizer with a learning rate of 0.001.
The results are reported in Tab.~\ref{tab:ablation_finetuning}. We observe that we outperform the baseline on all three settings.

\begin{table}[h]
\scriptsize
  \centering
      \caption{
    Performance comparison (\%) of ViT-Huge+TBM after fine-tuning on ImageNet-C.        
      }
      \begin{tabular}{cccc} 
	   \toprule 
      Methods & Supervised & Self-supervised & Fine-tuned \\
 	   \midrule  	   
 	   ViT-Huge \citeSupp{dosovitskiy2021an2} & 50.1 & 35.2 & 51.0  \\
	   ViT-Huge+TBM  & \textbf{53.2}  & \textbf{38.6}  & \textbf{54.1} \\ 	   
      \bottomrule
      \end{tabular}
	\label{tab:ablation_finetuning} 
\end{table}

\textbf{Comparison against alternative settings with similar model size.}
Here, we compare ViT-Huge+TBM against other settings with similar model size.
In \textbf{ViT-Huge+TBM ($\lambda=0$)}, we train our ViT-Huge+TBM model as usual, except that the weight of the TBM's reconstruction loss $\lambda$ is set to 0. 
This means that the TBM modules are still inserted into ViT-Huge, but there is no explicit boosting, and the TBM parameters are just a part of our model (which are involved in the end-to-end training).
Alternatively, another way to add more parameters to approximately match the number of parameters with our ViT-Huge+TBM model is to have slightly more layers (\textbf{ViT-Huge + more layers}) or slightly wider layers (\textbf{ViT-Huge + wider layers}).
As shown in Tab.~\ref{tab:ablation_model_capacity}, our method outperforms other methods with a similar model size.

\begin{table}[h]
  \centering
      \caption{
      Performance comparison against alternative settings with similar model size on ImageNet-C.
      ViT-Huge+TBM outperforms other variants with the same number of parameters.    
      }
      \begin{tabular}{cc} 
	   \toprule
      Settings & Accuracy (\%) \\
 	   \midrule
 	   ViT-Huge + more layers & 36.0 \\
 	   ViT-Huge + wider layers & 36.1 \\
 	   \midrule 
 	   ViT-Huge+TBM ($\lambda=0$) & 35.7 \\
 	   \midrule  	   
	   ViT-Huge+TBM  & 38.6 \\
      \bottomrule
      \end{tabular}
	\label{tab:ablation_model_capacity} 
\end{table}

\textbf{Impact of the design of $g$.}
Next, we investigate different designs for the module $g$, and the results are reported in Tab.~\ref{tab:g}.
Firstly, we ablate over the depth of $g$ on all three tasks, and find that when we increase the depth of $g$, the performance improves, and then keeps stable.
We find that using 3 layers for all tasks reaches optimal performance in the stable region.
Next, we also replace the 3 simple fc-layers with 3 self-attention layers, and find that the results are similar. Thus, overall, we use 3 fc-layers in our design.

\begin{table*}[ht]
    \caption{Evaluation of various designs of $g$ on three different tasks.}    
    \centering
    \begin{tabular}{c|c c c c c| c}    
      \toprule
      \# fc layers in $g$ & 1  &  2  & 3  & 4  & 5 & 3 self-attention \\ 
      \midrule
      Accuracy (\%) on ImageNet-C & 37.5 & 38.2 & 38.6 & 38.7 & 38.6 & 38.5 \\
      Accuracy (\%) on NTU60 & 78.6 & 79.0 & 79.1 & 79.1 & 79.0 & 79.0\\
      Accuracy (\%) on WRGBD & 71.5 & 71.4 & 71.5 & 71.4 & 71.5 & 71.4\\     
      \bottomrule
    \end{tabular}
    \label{tab:g}
    \vspace{-0.0cm}
\end{table*}

\section{More Implementation Details}

\subsection{Training on Images}

\textbf{Detailed image pre-training procedure.}
For the training on corrupted RGB and depth images, we employ the random sampling strategy as outlined in \citeSupp{he2021masked2}. Each image is split into 196 non-overlapping patches of approximately similar size. Then, patches are randomly selected to be masked until 75\% of them have been masked, i.e., 147 patches are masked, and 49 patches remain visible. When patches are masked, the corresponding tokens are \textit{removed} and are not fed as input to the VT encoder. Instead, masked tokens are introduced at the decoder for reconstruction purposes.
Following \citeSupp{he2021masked2}, we only apply MSE reconstruction loss on the masked patches, and do not apply it on visible patches.
During pre-training, following \citeSupp{he2021masked2}, our decoder consists of 8 stacked Transformer blocks with a width of 512. We also employ a data augmentation strategy of random resized cropping.

\textbf{Detailed image linear probe hyperparameters.}
After adding the linear layer, the VT encoder is frozen and the final layer is trained
using the cross entropy loss for 90 epochs with a batch size of 512. We use a LARS optimizer and set the learning rate to 0.1. We also augment our data using random resized cropping.

\textbf{Detailed image fully-supervised training hyperparameters.}
For the supervised setting on images, the full VT + linear layer is trained using the cross entropy loss for 400 epochs.
ViT backbones are optimized using LARS optimizer with a learning rate of 0.1, while DEiT and Swin backbones are optimized using AdamW optimizer with a learning rate of 0.001.
To augment our data, we use random resized cropping.

\subsection{Training on Skeleton Sequences}

\textbf{Detailed skeleton sequence tokenization procedure.}
Following DSTA-Net \citeSupp{shi2020decoupled2}, each input frame of the skeleton sequence is split into $P$ patches, where the $P$ presents the number of joints which is $25$ for NTU RGB+D datasets. 
Each patch is converted to a token embedding using a mapper comprising of a Conv2D layer with width of $256$, a BatchNorm layer and a LeakyReLU activation function.
We follow \citeSupp{liu2020disentangling2} to process all the skeleton sequences.

\textbf{Detailed skeleton sequence masking procedure.}
Here, we describe how we select the patches to mask during pre-training.
We follow the Spatial Temporal Masking strategy in \citeSupp{shan2022p}.
This strategy combines both Spatial Masking and Temporal Masking, where the model is forced to 1) learn to use relevant information from other (unmasked) joints, which improves the model's spatial awareness of the skeletal structure and 2) learn plausible motions of skeletal joints through an extended period of time, which improves the model's understanding of the temporal aspect of skeletal sequences.
Specifically, following \citeSupp{shan2022p}, we set the temporal masking ratio to 80\%, and spatial masking to involve 2 joints per frame.

\textbf{Detailed skeleton sequence pre-training procedure.}
When doing the masking procedure, we follow the same strategy in \citeSupp{shan2022p} where the masked joints are \textit{removed} and not fed to the following encoding process. Masked tokens are added before the skeleton decoder to be reconstructed into the original skeleton sequence.
For the skeleton decoder, we follow the basic structure of Kinetic-GAN \citeSupp{degardin2022generative2} that leverages $7$ stacked basic ST-GCN blocks as the decoder network. We train it for 200 epochs with batch size of 256.

\textbf{Detailed skeleton sequence linear probe hyperparameters.}
After adding the linear layer, the DSTA-Net encoder is frozen and the final layer is trained
using the cross entropy loss for 90 epochs with a batch size of 256. We use a SGD optimizer and set the learning rate to 0.1.

\textbf{Detailed skeleton sequence fully-supervised training hyperparameters.}
For the supervised setting on skeleton sequences, the DSTA-Net+linear layer is end-to-end trained
using the cross entropy loss for 90 epochs with a batch size of 256.
Here, we adopt an SGD optimizer with a learning rate of 0.1.

\subsection{More TBM Details}
In order to set $\alpha >0$ in practice while optimizing $\alpha$ using gradient descent, we apply the ReLU function onto our $\alpha$ parameters before using them to scale the noise.
The ReLU function will have the effect of mapping elements of $\alpha$ into values $ \geq 0$ for scaling the noise.
Moreover, we add a skip connection over the TBM module, which helps with convergence during training.

\section{Future Work}
In this work, we explore the scenario where training and testing corruption types are the same, which is commonly seen in many real-world tasks, such as skeleton action recognition and depth image recognition.
As shown in the main paper, our TBM can handle these tasks well and achieve state-of-the-art results.
In the future, we plan to explore the adaptation to unseen corruption types, i.e., Unsupervised Domain Adaptation (UDA) to new corruptions, which is an even more challenging scenario.
One possible solution is to modify TBM to generate $\alpha$ based on the input, which allows TBM to adapt flexibly according to the input data, i.e., similar to a dynamic network \citeSupp{foo2022era,li2022dynamic2} that adapts its parameters according to the input data.

Other future work includes experimentation on other tasks which also involve noisy input data, such as pose estimation \citeSupp{gong2023diffpose,foo2023system,zhao2019semantic,zhao2022graformer,xu2021graph,ng2022animal}. We can also investigate applying TBM to unified models \citeSupp{foo2023unified2,gupta2022towards,wang2022ofa,chen2022obj2seq} that tackle multiple tasks simultaneously.

\section{Analysis of $E [Q | I] =  E [P| I]$}

In Section 3.2 of our paper, we mention that $\mathbb{E} [Q | I] =  \mathbb{E} [P| I]$, given our assumptions that $P$ and $Q$ are drawn from the same distribution.
Here we give theoretical justifications of this.

\begin{align}
    \mathbb{E} [Q | I]  &=  \mathbb{E} [Q | R + P + Q = I] \\
    &= \int_q  q \cdot \frac{\text{Prob}(Q=q, R + P + Q = I)}{\text{Prob}(R + P + Q = I)} \\
    &= \int_q  q \cdot \frac{\text{Prob}(Q=q, P = I - R - q)}{\text{Prob}(R + P + Q = I)} \\
    \label{eq:Q_side}    
    &= \frac{1}{d} \int_q  q \cdot \text{Prob}(Q=q, P = I - R - q)
\end{align}
where $d = \text{Prob}(R + P + Q = I)$ is a constant.

On the other hand, we also get:

\begin{align}
    \mathbb{E} [P | I]  &=  \mathbb{E} [P | R + P + Q = I] \\
    &= \int_p  p \cdot \frac{\text{Prob}(P=p, R + P + Q = I)}{\text{Prob}(R + P + Q = I)} \\
    &= \int_p  p \cdot \frac{\text{Prob}(P=p, Q = I - R - p)}{\text{Prob}(R + P + Q = I)}  \\
    \label{eq:P_side}
    &= \frac{1}{d} \int_p  p \cdot \text{Prob}(P=p, Q = I - R - p)    
\end{align}

When $P$ and $Q$ have the same distribution, and thus the same support, we can conclude that Eq.~\ref{eq:Q_side} =  Eq.~\ref{eq:P_side}, and $\mathbb{E} [Q | I] =  \mathbb{E} [P| I]$.

Note that above, we have the assumption that $P$ and $Q$ have the same distribution. Thus, in the next section, we analyse theoretically why our method can learn to enable the synthetic corruptions $Q$ to have the same distribution as $P$.

\section{Analysis that distribution of synthetic corruptions $Q$ approximate distribution of natural corruptions $P$}

In the section above, we assume that $Q$ and $P$ come from the same distribution. Here, we show why we can expect this in practice, and show theoretical analysis to support this.
In short, it is because $\alpha$ can be meaningfully trained to be similar to the underlying corruption distribution. Specifically, the achieved loss will be higher, if $\alpha$ does not learn to model the underlying corruption distribution well.

Following \citeSupp{shi2019probabilistic2,li2019unsupervised2,yu2019robust2}, the ``natural" corruption distribution $\mathcal{P}$ is modeled as Gaussian with a mean of 0 and some unknown standard deviations $\omega$.
In our TBM, the synthetic corruption $Q$ is drawn from Gaussian with standard deviations $\alpha$. 
We next show that if $\alpha \neq \omega$, the produced boosted features $\hat{R}$ will be biased, and so there will be a higher loss incurred.

To show that $\mathbb{E} [Q | I] = \frac{\alpha^2}{\omega^2} \mathbb{E} [P | I]$, we
first start by analyzing the conditional pdf of $P$.

\begin{align}
    &\text{Prob}(P=p | I=i, R=r) = \frac{\text{Prob}(P=p, I=i, R=r) }{ \text{Prob}(I=i, R=r) } \\ 
    \label{eq:split_marginals}
    &= \frac{\text{Prob}(P=p)\text{Prob}(Q= i-r-p) }{ \text{Prob}(I=i, R=r) } \\    
    &= \frac{1}{b_1} \text{Prob}(P=p) \text{Prob}(Q= i-r-p) \\
    &= \frac{1}{b_2} exp \left\{ -\frac{p^2}{2 \omega^2}   \right\} exp \left\{ -\frac{(i - r - p)^2}{2 \alpha^2}   \right\} \\    
    &= \frac{1}{b_2 } exp \left\{ -\frac{\alpha^2 p^2 + \omega^2(i - r - p)^2}{2 \alpha^2 \omega^2}   \right\}  \\
    &= \frac{1}{b_2} exp \left\{ -\frac{(\alpha^2 + \omega^2) p^2 - 2\omega^2 p(i - r) + \omega^2 (i-r)^2}{2 \alpha^2 \omega^2}   \right\} \\     
    &= \frac{1}{b_2} exp \left\{ -\frac{ p^2 - \frac{2\omega^2}{(\alpha^2 + \omega^2)} p(i - r) + \frac{\omega^2}{(\alpha^2 + \omega^2)} (i-r)^2}{2 \frac{ \alpha^2 \omega^2}{(\alpha^2 + \omega^2)}}   \right\} \\
    &= \frac{1}{b_2} exp \left\{ -\frac{ p^2 - \frac{2\omega^2}{(\alpha^2 + \omega^2)} p(i - r) + \big(\frac{\omega^2}{(\alpha^2 + \omega^2)} \big)^2 (i-r)^2}{2 \frac{ \alpha^2 \omega^2}{(\alpha^2 + \omega^2)}}   \right\}  \nonumber \\         
    &~~~~\cdot exp \left\{ -\frac{\frac{\omega^2}{(\alpha^2 + \omega^2)} (i-r)^2 - \big(\frac{\omega^2}{(\alpha^2 + \omega^2)} \big)^2 (i-r)^2}{2 \frac{ \alpha^2 \omega^2}{(\alpha^2 + \omega^2)} }   \right\} \\    
    \label{eq:merged_conditional_dist}
    &= \frac{1}{b_3 } exp \left\{ -\frac{ (\frac{\omega^2}{\alpha^2 + \omega^2} (i-r) - p )^2 }{2 \frac{ \alpha^2 \omega^2}{(\alpha^2 + \omega^2)}}   \right\}     
\end{align}
where $b_1,b_2,b_3$ are constants.
To get Eq.~\ref{eq:split_marginals} by splitting the joint distribution into marginals, we use the fact that $I = R + P + Q$, and $P$ and $Q$ are independent of each other.
We note that the conditional distribution in Eq. \ref{eq:merged_conditional_dist} has a mean of $\frac{\omega^2}{\alpha^2 + \omega^2} (i-r)$. In other words, $\mathbb{E} [P | I = i, R=r] = \frac{\omega^2}{\alpha^2 + \omega^2} (i-r)$.
Doing the same for the conditional distribution of $Q$ gives us $\mathbb{E} [Q | I=i, R=r] = \frac{\alpha^2}{\alpha^2 + \omega^2} (i-r)$.
Equating both of them gives us the resulting relationship: $\mathbb{E} [Q| I = i, R = r] = \frac{\alpha^2}{\omega^2} \mathbb{E} [P| I = i, R = r]$.
As this result holds independently of $R$,
we get $\mathbb{E} [Q|I] = \frac{\alpha^2}{\omega^2} \mathbb{E} [P|I]$.

In our TBM module, we will use Eq. 1 of the main paper, i.e., $\hat{R} = 2 \hat{F} - I$ to reconstruct $\hat{R}$. Thus, our estimate becomes:

\begin{align}
    &\hat{R} = 2 \mathbb{E} [R + P|I] - (\mathbb{E} [R+P+Q|I]   \\
     &= 2 \mathbb{E} [R|I] + 2 \mathbb{E} [P|I] - (\mathbb{E} [R|I] + \mathbb{E} [P|I] + \mathbb{E} [Q|I])   \\
     &= \mathbb{E} [R|I]  + \mathbb{E} [P|I] - \mathbb{E} [Q|I]  \\
     &=  \mathbb{E} [R|I]  +  \mathbb{E} [P|I]  -  \frac{\alpha^2}{\omega^2} \mathbb{E} [P|I] \\
     &= \mathbb{E} [R|I]  +  ( 1 - \frac{ \alpha^2}{\omega^2}) \mathbb{E} [P|I] 
     \label{eq:diff_noise_R_recon}
\end{align}
If $\alpha \neq \omega$, this obtained term in Eq.~\ref{eq:diff_noise_R_recon} is not equal to $ \mathbb{E} [R|I]$, which is not desirable as it means that our boosted features will be biased, i.e., $\hat{R} \neq \mathbb{E} [R|I]$.
For example, when $\alpha << \omega$ and $\frac{\alpha^2}{\omega^2}$ is small and close to 0, we get an estimate $\hat{R} \approx \mathbb{E} [R|I]  +  \mathbb{E} [P|I] = F $, which means that almost no boosting is done.
On the other hand, when $\alpha >> \omega$, our TBM module will over-compensate for the corruptions, and we get a case where the corruption changes signs (from $ \mathbb{E} [P|I]$ to $ ( 1 - \frac{ \alpha^2}{\omega^2}) \mathbb{E} [P|I]$) and still affect the performance of the task.

Failure to boost the tokens in the TBM module will lead to a higher loss in the end-to-end objective, as analyzed in Sec.~\ref{sec:appendix_analysis} of the Supplementary. 
Thus, to minimize this loss, the gradients will optimize $\alpha$ to become an approximation of $\omega$, i.e., $\alpha \approx \omega$. Thus, $\frac{\alpha^2}{\omega^2} \approx 1$ and Eq.~\ref{eq:diff_noise_R_recon} becomes $\mathbb{E} [R|I] + ( 1 - \frac{ \alpha^2}{\omega^2}) \mathbb{E} [P|I] \approx \mathbb{E} [R|I]$ as we intend, in order to minimize the loss.
This means that $\alpha$ will be trained to be similar to $\omega$, the parameters of the ``natural" corruption distribution.

\section{Detailed Analysis of Section 3.4 of Main Paper}
\label{sec:appendix_analysis}

The detailed derivation of Eq.~5 in the main paper is as follows:
\begin{align}
    \mathbb{E} &\bigg[ \frac{1}{N_V} \sum_{j=1}^{N_V} [V_j - (\beta U + c)_j]^2 \bigg] \\
    &= \mathbb{E} \bigg[  \frac{1}{N_V} \sum_{j=1}^{N_V} [(\beta U + c + \epsilon)_j - (\beta U + c)_j]^2  \bigg] \\
    &= \mathbb{E} \big[  \frac{1}{N_V} \sum_{j=1}^{N_V} \epsilon_j^2  \big]\\
    &= \frac{1}{N_V} \sum_{j=1}^{N_V} \mathbb{E} [\epsilon_j^2]  \\
    &= \frac{1}{N_V} \sum_{j=1}^{N_V} Var(\epsilon_j) + \mathbb{E} [\epsilon_j]^2 \\
    &= \gamma^2
\end{align}

Next, the detailed steps to get Eq.~6-8 of the main paper is as follows:
\begin{align}
    \mathbb{E} &\bigg[ \frac{1}{N_V} \sum_{j=1}^{N_V} [\tilde{V}_j - (\beta \tilde{U} + c)_j]^2 \bigg] \\
    &= \mathbb{E} \bigg[ \frac{1}{N_V} \sum_{j=1}^{N_V} [(V + S_V)_j - (\beta (U+S_U) + c)_j]^2 \bigg] \\
    &= \mathbb{E} \bigg[ \frac{1}{N_V} \sum_{j=1}^{N_V} [\epsilon_j + (S_V)_j - (\beta S_U )_j]^2 \bigg] \\  
    &= \frac{1}{N_V} \sum_{j=1}^{N_V} \mathbb{E} \bigg[  [\epsilon_j]^2 + [(S_V)_j]^2 +  [(\beta S_U )_j]^2
    \nonumber \\
    \label{eq:end_of_first_part}
    &~~~ - 2[(S_V)_j][(\beta S_U )_j]  - 2[\epsilon_j][(\beta S_U )_j] + 2[\epsilon_j][(S_V)_j] \bigg] \\
    &= \frac{1}{N_V} \sum_{j=1}^{N_V} \mathbb{E} [\epsilon_j^2] +  \mathbb{E} [(S_V)_j^2] + \mathbb{E} [(\beta S_U)_j^2 ]   \nonumber \\
    &~~~~ -2\mathbb{E} [(S_V)_j \cdot (\beta S_U )_j] - 2\mathbb{E} [\epsilon_j \cdot (\beta S_U )_j] \nonumber \\
    &~~~~ + 2\mathbb{E} [\epsilon_j \cdot (S_V)_j]  \\    
    &= \frac{1}{N_V} \sum_{j=1}^{N_V} \mathbb{E} [\epsilon_j^2] + \mathbb{E} [(S_V)_j^2] +  \mathbb{E} [(\beta S_U)_j^2 ]  \nonumber \\
    &~~~~ - 2\mathbb{E} [\epsilon_j]\mathbb{E}[ (\beta S_U )_j] -2\mathbb{E} [(S_V)_j] \mathbb{E}[(\beta S_U )_j]  \nonumber \\    
    &~~~~ + 2\mathbb{E} [\epsilon_j]\mathbb{E}[ (S_V)_j] ~~\text{(By independence)} \\        
    &= \frac{1}{N_V} \sum_{j=1}^{N_V} \mathbb{E} [\epsilon_j^2] + \mathbb{E} [(S_V)_j^2] +  \mathbb{E} [(\beta S_U)_j^2 ] \nonumber \\
    &~~~~~\text{(as $\mathbb{E} [\epsilon_j] = \mathbb{E} [(S_V)_j] =0$)}  \label{eqn:split_before} \\
    &= \frac{1}{N_V} \sum_{j=1}^{N_V} (Var(\epsilon_j) + \mathbb{E} [\epsilon_j]^2) + (Var((S_V)_j)    \nonumber \\
    &~~~~ + \mathbb{E} [(S_V)_j]^2) + (Var((\beta S_U)_j) + \mathbb{E} [(\beta S_U)_j]^2) \\      
    &= \frac{1}{N_V} \sum_{j=1}^{N_V} Var(\epsilon_j)  + Var((S_V)_j) \nonumber \\
    &~~~~ + Var((\beta S_U)_j)  ~~\text{(By independence)} \\
     &= \frac{1}{N_V}  \sum_{j=1}^{N_V} \bigg( \gamma^2 + \sigma_N^2 + \sum_{k=1}^{N_U} Var(\beta_{jk} (S_{U})_k) \bigg) \\    
    &= \frac{1}{N_V} \sum_{j=1}^{N_V} \gamma^2 + \sigma_N^2 + \sum_{k=1}^{N_U} \beta_{jk}^2 \sigma_N^2  \\
    \label{noisy_loss}    
    &= \gamma^2 + \sigma_N^2 + \frac{1}{N_V} \sum_{j=1}^{N_V} \sum_{k=1}^{N_U} \beta_{jk}^2 \sigma_N^2 
\end{align}

Lastly, the detailed steps corresponding to Eq.~9-11 of the main paper are as follows:
\begin{align}
    \mathbb{E} & \bigg[\frac{1}{N_V} \sum_{j=1}^{N_V} [\tilde{V}_j - (\beta U + c)_j]^2 \bigg] \\
    &= \mathbb{E} \bigg[ \frac{1}{N_V} \sum_{j=1}^{N_V} [(V + S_V)_j - (\beta U + c)_j]^2  \bigg] \\
    &= \mathbb{E} \bigg[ \frac{1}{N_V} \sum_{j=1}^{N_V} [\epsilon_j + (S_V)_j ]^2  \bigg] \\  
    &= \mathbb{E} \bigg[ \frac{1}{N_V} \sum_{j=1}^{N_V} [\epsilon_j]^2  + [(S_V)_j]^2 +  2[\epsilon_j][(S_V)_j] \bigg] \\  
    &= \frac{1}{N_V} \sum_{j=1}^{N_V} \mathbb{E} [\epsilon_j^2] + \mathbb{E} [(S_V)_j^2] +  2\mathbb{E} [\epsilon_j (S_V)_j] \\    
    &= \frac{1}{N_V} \sum_{j=1}^{N_V} \mathbb{E} [\epsilon_j^2]  +  \mathbb{E} [(S_V)_j^2] + 2\mathbb{E} [\epsilon_j]\mathbb{E}[ (S_V)_j] \\
    \nonumber
    &~~\text{(By independence)} \\        
    &= \frac{1}{N_V} \sum_{j=1}^{N_V} \mathbb{E} [\epsilon_j^2]  + \mathbb{E} [(S_V)_j^2]  ~~\text{(as $\mathbb{E} [\epsilon_j] = \mathbb{E} [(S_V)_j] =0$)} \\    
    &= \frac{1}{N_V} \sum_{j=1}^{N_V} (Var(\epsilon_j) + \mathbb{E} [\epsilon_j]^2) + (Var((S_V)_j) + \mathbb{E} [(S_V)_j]^2)  \\
    &= \frac{1}{N_V} \sum_{j=1}^{N_V} Var(\epsilon_j) + Var((S_V)_j)  \\
    &= \gamma^2 + \sigma_N^2     
\end{align}

\clearpage

{\small
\bibliographystyleSupp{ieee_fullname}
\bibliographySupp{egbib_supp}
}

\end{document}